%% file: 0-main.tex
\documentclass[conference]{IEEEtran}
\IEEEoverridecommandlockouts
\usepackage{cite}
\usepackage{amsmath,amssymb,amsfonts}
\usepackage{algorithmic}
\usepackage{graphicx}
\usepackage{textcomp}
\usepackage{xcolor}

\usepackage{booktabs}
\usepackage{algorithm}
\usepackage{mathrsfs}
\usepackage{makecell}
\usepackage{enumitem}

\def\BibTeX{{\rm B\kern-.05em{\sc i\kern-.025em b}\kern-.08em
    T\kern-.1667em\lower.7ex\hbox{E}\kern-.125emX}}

\newcommand{\framework}{SLED}
    
\begin{document}

\title{Unsupervised Skin Lesion Segmentation via Structural Entropy Minimization \\ 
on Multi-Scale Superpixel Graphs}

\author{
\IEEEauthorblockN{Guangjie Zeng\IEEEauthorrefmark{1}, Hao Peng\IEEEauthorrefmark{1}$^\#$\thanks{$^\#$Corresponding author.}, Angsheng Li\IEEEauthorrefmark{1}\IEEEauthorrefmark{2}, Zhiwei Liu\IEEEauthorrefmark{3}, Chunyang Liu\IEEEauthorrefmark{4}, Philip S. Yu\IEEEauthorrefmark{5}, Lifang He\IEEEauthorrefmark{6}}
% \IEEEauthorblockA{\IEEEauthorrefmark{1}Beihang University; \IEEEauthorrefmark{2}Zhongguancun Laboratory; \IEEEauthorrefmark{3}Salesforce AI Research; \IEEEauthorrefmark{4}Didi Chuxing; \\ \IEEEauthorrefmark{5}University of Illinois at Chicago; \IEEEauthorrefmark{6}Lehigh University. \\
% \{zengguangjie, penghao, angsheng\}@buaa.edu.cn, zhiweiliu@salesforce.com, \\ liuchunyang@didiglobal.com, psyu@uic.edu, lih319@lehigh.edu.
\IEEEauthorblockA{\IEEEauthorrefmark{1}\textit{State Key Laboratory of Software Development Environment, Beihang University}, Beijing, China; \\ \IEEEauthorrefmark{2}\textit{Zhongguancun Laboratory}, Beijing, China; \IEEEauthorrefmark{3}\textit{Salesforce AI Research}, California, USA; \\  \IEEEauthorrefmark{4}\textit{Didi Chuxing}, Beijing, China;  \IEEEauthorrefmark{5}\textit{University of Illinois Chicago}, Illinois, USA; \IEEEauthorrefmark{6}\textit{Lehigh University}, Pennsylvania, USA. \\
\{zengguangjie, penghao, angsheng\}@buaa.edu.cn, zhiweiliu@salesforce.com, \\ liuchunyang@didiglobal.com, psyu@uic.edu, lih319@lehigh.edu.
}
}

\maketitle

\begin{abstract}
Skin lesion segmentation is a fundamental task in dermoscopic image analysis.
The complex features of pixels in the lesion region impede the lesion segmentation accuracy, and existing deep learning-based methods often lack interpretability to this problem.
In this work, we propose a novel unsupervised \underline{S}kin \underline{L}esion s\underline{E}gmentation framework based on structural entropy and isolation forest outlier \underline{D}etection, namely~\framework.
Specifically, skin lesions are segmented by minimizing the structural entropy of a superpixel graph constructed from the dermoscopic image.
Then, we characterize the consistency of healthy skin features and devise a novel multi-scale segmentation mechanism by outlier detection, which enhances the segmentation accuracy by leveraging the superpixel features from multiple scales. 
We conduct experiments on four skin lesion benchmarks and compare \framework~with nine representative unsupervised segmentation methods.
Experimental results demonstrate the superiority of the proposed framework. 
Additionally, some case studies are analyzed to demonstrate the effectiveness of \framework.
\end{abstract}


\input{1-Introduction}
\input{2-Related}
\input{3-Method}
\input{4-Experiments}

\input{5-Conclusion}

\section*{Acknowledgement}
This work is supported by the National Key R\&D Program of China through grant 2021YFB1714800, NSFC through grant 61932002, S\&T Program of Hebei through grant 20310101D, Natural Science Foundation of Beijing Municipality through grant 4222030, CCF-DiDi GAIA Collaborative Research Funds for Young Scholars, and the Fundamental Research Funds for the Central Universities. 
Philip S. Yu is supported by NSF under grant III-2106758.
\bibliographystyle{IEEEtran}
\bibliography{references}

\end{document}

%% file: 1-Introduction.tex
\section{Introduction}\label{sec:intro}
Skin lesion segmentation of dermoscopic images is essential for skin cancer diagnosis \cite{yuan2017automatic}.
Accurate segmentation of skin lesions from the surrounding skin is challenging because skin lesions, especially melanoma ones, vary in color, texture, shape, size, and location.
Moreover, low contrast between the lesion and healthy skin, changing illumination, and the existence of artifacts like hairs, blood vessels, and color calibration charts make this task even harder.

\begin{figure}[t]
    \centering
    \includegraphics[width=1\linewidth]{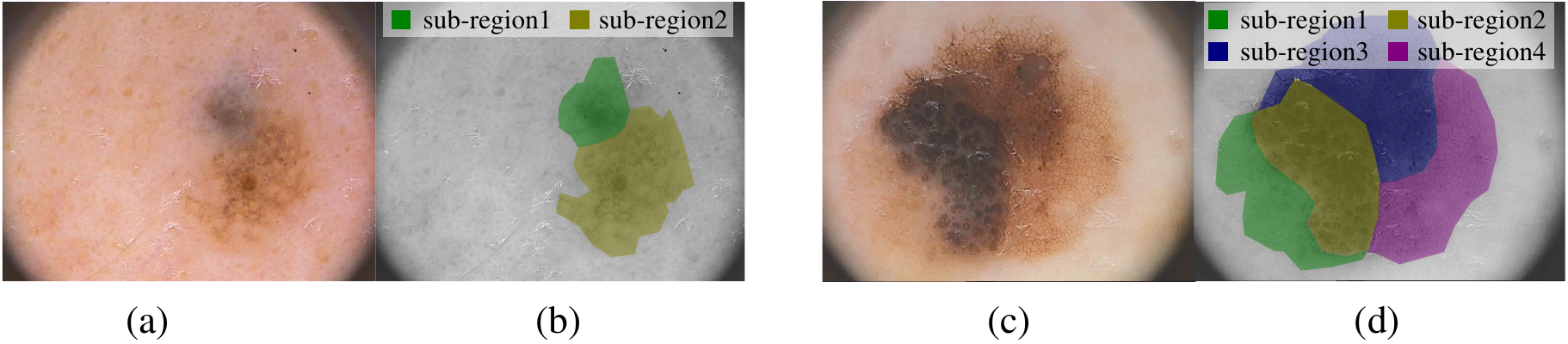}
    \caption{
    % Examples of skin lesion dermoscopic images. (a) A benign lesion; (b)the segmentation regions of (a), which contains two sub-regions; (c) a melanoma lesion; (d) the segmentation regions of (c), which contains four sub-regions.
    Examples of skin lesion dermoscopic images. (a) Example lesion 1; (b) The segmentation regions of (a), which contains two sub-regions; (c) Example lesion 2; (d) The segmentation regions of (c), which contains four sub-regions.
    }
    \label{fig:challenge}
\end{figure}

A large amount of unsupervised skin lesion segmentation methods have been proposed by researchers in recent years, including thresholding-based methods \cite{guarracino2018sdi+}, region-based methods \cite{patino2018automatic}, saliency-based methods \cite{zhou2023saliency}, and clustering-based methods \cite{ahn2021spatial}.
They segment skin lesions without labeled datasets, relieving medical experts from the laborious work of data labeling.
% However, the fine-grained characteristics of different types of lesion skins and healthy skins are under-explored. 
However, the fine-grained characteristics of lesion skins and healthy skins are under-explored.
% Figure \ref{fig:challenge} illustrates examples of benign and melanoma lesions from the PH2 dataset \cite{mendoncca2013ph}.
% On the one hand, the features like color and texture of pixels in the lesion regions are diverse, especially for melanoma lesions which present uneven color distributions \cite{abbasi2004early}, as compared in Figure~\ref{fig:challenge}. 
Figure \ref{fig:challenge} illustrates examples of skin lesions from the PH2 dataset \cite{mendoncca2013ph}.
On the one hand, the features like color and texture of pixels within the lesion regions are diverse, and the lesion region can be further divided into some unspecified number of sub-regions as shown in Figures \ref{fig:challenge}(b) and \ref{fig:challenge}(d).
Ignoring the diversity in a fine-grained level of those pixels leads to sub-optimal segmentation results.  
On the other hand, the features of pixels in healthy skin are generally consistent, which is one of the most important prior information for skin lesion segmentation. 
Existing methods fail to fully and effectively utilize this prior, thus being unable to accurately segment the margin between the lesion region and healthy skin.

Regarding the diversity of pixel features in the lesion region, some works directly segment the dermoscopic image into two regions, \textit{i.e.}, the lesion region and healthy skin \cite{guarracino2018sdi+,patino2018automatic}.
They ignore the fine-grained level distinction within the lesion region, 
which can lead to inaccurate segmentation of pixels at lesion borders.
Other works segment dermoscopic images into a proper number of regions \cite{ahn2021spatial}, and then classify regions into the lesion region and healthy skin.
Though being better at distinguishing the fine-grained features, they require a pre-defined number of sub-regions, which lacks generalization ability to new scenarios.
Regarding the consistent healthy skin prior, RSSLS \cite{ahn2017saliency} utilizes reconstruction error-based saliency to segment skin lesions.
With the detected background, \textit{i.e.}, the healthy skin, the whole image is reconstructed based on the background, forming the saliency map.
However, the background detection process is complex and imprecise, inducing noise into the segmentation procedure.

In this work, we propose a novel unsupervised \underline{S}kin \underline{L}esion s\underline{E}gmentation framework via structural entropy~(SE) and outlier \underline{D}etection, namely~\framework, which adaptively determines the number of segments by SE minimization, and utilizes the consistent healthy skin prior through isolation forest outlier detection.
The basic idea is that we take image segmentation as the graph partition problem where graphs are constructed to incorporate both color features and spatial position of pixels (or superpixels).
First, we segment the dermoscopic image into a proper number of regions by minimizing the SE of a graph constructed from the image without the need to know the number of sub-regions a priori, which is critical as shown in Figure \ref{fig:challenge} the number of sub-regions of a lesion is unknown and unpredictable.  
% First, we segment the dermoscopic image into the proper number of regions by minimizing the SE of a K-Nearest Neighbor (K-NN) graph constructed from the image, which also has the advantage of easily keeping the spatial continuity of segments.
% to answer the question of why should we use graph-based methods for image segmentation
The nodes of the graph are pre-segmented superpixels and edge weights are the similarity between superpixels.
Modules in the graph partition obtained by SE minimization represent the regions.
Second, we distinguish these regions into the lesion region and healthy skin by maximizing the between-class variance, forming a single-scale segmentation.
Third, we integrate single-scale segmentation results from different superpixel scales into multi-scale segmentation.
To achieve this integration, we train isolation forest models using superpixels in the healthy skin regions from single-scale segmentation results and generate a multi-scale outlier score map by combining outlier scores from different scales.
The consistency of pixel features in healthy skin helps isolation forest models to distinguish the lesion region from healthy skin with higher outlier scores.
Thresholding on the multi-scale score map gives the multi-scale segmentation.
Source code is available on Github\footnote{https://github.com/SELGroup/SLED}.

We comprehensively evaluate our framework with state-of-the-art unsupervised segmentation methods and several clustering-based methods on four popular skin lesion segmentation datasets.
The results show that our framework is more accurate in skin lesion segmentation.
We also test the performance of several outlier detection methods, finding that isolation forest best models healthy skin in dermoscopic images.
The main contributions of this paper are summarized as follows: 
(1) An unsupervised skin lesion segmentation framework guided by iteratively refined structural entropy is proposed, which adaptively determines the number of segments in the segmentation process.
(2) We characterize the fact that features in healthy skin are consistent and devise a novel multi-scale segmentation mechanism via isolation forest outlier detection method. 
(3) Comprehensive experiments on four popular skin lesion segmentation datasets demonstrate that our framework achieves new state-of-the-art performance among unsupervised skin lesion segmentation methods.

%% file: 2-Related.tex
\section{Related Work}
% In this section, we first give a brief review of unsupervised skin lesion segmentation, then we introduce several unsupervised image segmentation research lines related to skin lesion segmentation, and finally give a short introduction to SE.

\subsection{Unsupervised Skin Lesion Segmentation}
Compared to supervised methods, unsupervised skin lesion segmentation methods avoid the laborious work of lesion labeling, and have attracted great interest in recent years.
Here we briefly review existing unsupervised skin lesion segmentation methods, including thresholding-based, region-based, saliency-based, and clustering-based approaches.

Histogram thresholding is a simple yet effective way for skin lesion segmentation.
The thresholding histograms are constructed in different ways, such as principle component analysis (PCA) on color bands \cite{peruch2013simpler} and carefully chosen color bands \cite{guarracino2018sdi+}.
More specialized thresholding techniques are also designed for skin lesion segmentation, including weighted thresholding \cite{zortea2017simple} and adaptive thresholding \cite{thanh2019skin}.
Region-based methods iteratively merge image regions until the desired lesion region is found.
In \cite{patino2018automatic}, a simple region merging method that merges superpixels into two regions, i.e., the lesion region and healthy skin, is proposed.
In \cite{navarro2018accurate}, a superpixel region labeling method via continuity classification for skin lesion segmentation is proposed.
Saliency detection-based methods treat skin lesions in dermoscopic images as the salient object.
They construct a saliency map by assigning each pixel a saliency score, and segment the lesion region by thresholding on the saliency map.
The saliency scores are defined by reconstruction errors \cite{ahn2017saliency}, color and brightness prior \cite{fan2017automatic}, color channel volume\cite{zhou2023saliency} in unsupervised manners, or lists of regional contrast and regional property in a supervised manner \cite{jahanifar2018supervised}.
 Clustering-based methods group pixels or superpixels into regions and then classify them to find the skin lesion.
Ahn \textit{et al.}\cite{ahn2021spatial} proposed a spatial guided self-supervised clustering network (SGSCN) for skin lesion and liver tumor segmentation that iteratively determines the number of segments.
Zhang \textit{et al.}\cite{zhang2023deep} proposed a deep hyperspherical clustering method based on the theory of belief functions to achieve unsupervised skin lesion segmentation.
In the single-scale segmentation step of our framework, we group superpixels by minimizing SE and classify regions by maximizing the between-class variance.

\subsection{Unsupervised Image Segmentation}
Apart from methods specifically designed for skin lesion segmentation, there exist a large number of unsupervised image segmentation methods for more general usage.
Classical image segmentation methods group pixels into meaningful regions. 
Representative methods include $K$-means and spectral clustering \cite{shi2000normalized}.
More recent works utilize the power of deep learning such as convolutional neural networks (CNNs) to learn vision features.
Kanezaki \textit{et al.} \cite{kim2020unsupervised} proposed a CNN-based model to group spatially continuous pixels of similar features to facilitate unsupervised natural image segmentation. 
In order to perform the segmentation of the target region, there are several other research lines in unsupervised image segmentation related to skin lesion segmentation, which focuses on different scenarios.
The first line is unsupervised salient object detection, which aims at localizing and segmenting salient objects simultaneously without manual annotations.
Yan \textit{et al.} \cite{yan2022unsupervised} pointed out the unreliability of usually used noisy labels generated by traditional unsupervised salient object detection methods and proposed unsupervised domain adaptive salient object detection which learns from synthetic labels with higher quality.
Zhou \textit{et al.} \cite{zhou2023texture} proposed a texture-guided saliency distilling unsupervised salient object detection model through a confidence-aware saliency distilling strategy that assigns scores to samples based on confidence, and a boundary-aware texture matching strategy which refines boundaries by matching textures.
The second line is unsupervised foreground extraction, which aims for a binary partition of an image with a foreground containing identifiable objects and a background with remaining regions.
Yu \textit{et al.} \cite{yu2021unsupervised} presented deep region competition (DRC) to solve the foreground extraction problem by reconciling energy-based prior with generative image modeling in the form of mixture of experts.
Ding \textit{et al.} \cite{ding2022comgan} pointed out two kinds of trivial solutions in the image compositional generation process when performing foreground extraction, and solved trivial solutions with the proposed ComGAN model.
For the task of skin lesion segmentation, lesions can be taken as salient objects or image foregrounds, so both methods belonging to these two research lines can be used to tackle the task of skin lesion segmentation.

\subsection{Structural Entropy}\label{sec:se}
SE \cite{li2016structural} is first proposed to measure the complexity of the hierarchical structure of graphs through associated encoding trees, which is naturally a node clustering method.
As the SE of a graph $G$ is the minimum overall number of bits required to determine the codewords of the graph nodes, it measures the uncertainty embedded in $G$.
SE has been successfully applied in the field of bioinformatics \cite{li2018decoding}, reinforcement learning~\cite{zeng2023effective,zeng2023hierarchical}, and graph neural networks \cite{wu2022structural,zou2023se,yang2023minimum}.
In this work, we use the two-dimensional SE, \textit{i.e.}, SE of a graph by a partition, to perform skin lesion segmentation.
Without confusion, SE in the following sections means two-dimensional SE.

Let $G=(V,E,\mathbf{W})$ be an undirected weighted graph, where $V=\{v_1,...,v_N\}$ is the node set, $E$ is the edge set, and $\mathbf{W} \in \mathbb{R}^{N\times N}$ is the edge weight matrix.
A graph partition $\mathcal{P}$ defined on $G$ can be formulated as $\mathcal{P}=\{X_1,X_2,...,X_L\}$, where $X_i$ is the $i$-th module containing a subset nodes of $V$.
Given a graph $G$ and its graph partition $\mathcal{P}$, the SE of $G$ given by $\mathcal{P}$ is defined as:
\begin{equation}
\begin{aligned}
    \mathcal{H}^\mathcal{P}(G) = -\sum_{X \in \mathcal{P}} \sum_{v_i \in X}\frac{g_i}{vol(G)} {\rm log_2}\frac{d_i}{vol(X)} \\ 
    -\sum_{X \in \mathcal{P}}\frac{cut(X)}{vol(G)} {\rm log_2}\frac{vol(X)}{vol(G)},
\end{aligned}
\end{equation}
where $d_i$ is the degree of node $v_i$, $g_i$ is the weight sum of edges connecting $v_i$ and other nodes, $vol(X)$ and $vol(G)$ are volumes, \textit{i.e.}, the sum of the node degrees, in module $X$ and graph $G$, respectively, and $cut(X)$ is the cut of $X$, \textit{i.e.}, the weight sum of edges between nodes in and not in module $X$.
% where $d_i$ is the degree of node $i$, $vol(X)$ and $vol(G)$ are volumes, \textit{i.e.}, the sum of the node degrees, in module $X$ and graph $G$, respectively, and $g_i$, $cut(X)$ are the cut, i.e., the weight sum of edges with one endpoint in, of node $i$, module $X$ respectively \cite{li2016structural}.

%% file: 3-Method.tex
\section{Methodology}\label{sec:method}
\begin{figure*}[htbp]
\centering
\includegraphics[width=0.9\linewidth]{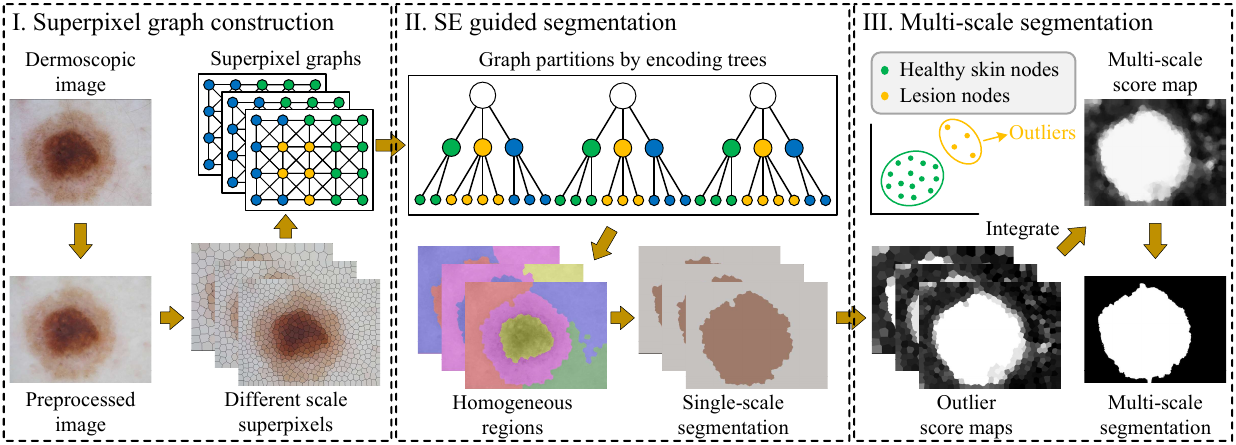}
\caption{Overview of \framework.}
\label{fig:framework}
\end{figure*}

In this section, we present \framework, a novel unsupervised skin lesion segmentation framework based on SE and outlier detection.
The overview of~\framework~is depicted in Figure \ref{fig:framework}.
\framework~composes of three components: superpixel graph construction, SE guided segmentation, and multi-scale segmentation mechanism by outlier detection.
Different scale superpixels are grouped by SE to form single-scale segmentation results, which are then integrated by the outlier detection-based multi-scale mechanism to generate the final segmentation.

\subsection{Basic Notations}
In this paper, superpixel graphs are constructed from images to perform segmentation, so some notations refer to elements both in graphs and images.
In particular, $v$ denotes both a graph node and a superpixel, $\mathcal{P}$ denotes both a graph partition and an image segmentation, and $X$ and $Y$ denote both modules in graph partition and regions in images. We list some important notations in Table \ref{tab:notations}.
\begin{table}[t]
\caption{Definition of notations.\label{tab:notations}}
\begin{center}
\begin{tabular}{@{}lcl@{}} \toprule
Symbol & Domain & Description  \\ \midrule
$G$ & - & Superpixel graph \\ 
$V, E, \mathbf{W}$ & - & Node set, edge set, edge weight matrix \\
$v$ & - & graph node/superpixel \\
% $N$ & $\mathbb{Z}$ & Number of graph nodes/superpixels \\
% $i,j$ & $\mathbb{Z}$ & Index of graph nodes/superpixels \\
$\mathcal{P}$ & - & Graph partition (image segmentation) \\ 
$L$ & $\mathbb{Z}$ & Number of modules/regions \\ 
$X, Y$ & - & Modules in graph partition (regions in images) \\ 
$\mathcal{H}^\mathcal{P}(G)$ & - & SE of $G$ given by $\mathcal{P}$ \\
$cut(X)$ & $\mathbb{R}$ & Cut of $X$ \\ &&(weight sum of edges connected to nodes in $X$) \\
$vol(X)$ & $\mathbb{R}$ & Volume of $X$ (degree sum of nodes in $X$) \\
% $\omega_{X}$ & $\mathbb{R}$ & Occurrence of region $X$ \\
% $\mu_{X}$ & $\mathbb{R}$ & Mean intensity of region $X$ \\
$\sigma^2_B$ & $\mathbb{Z}$ & Between class variance of regions \\
$o$ & $\mathbb{R}$ & Outlier score of a superpixel \\
$O$ & $\mathbb{Z}^{N \times N}$ & Outlier score map \\
$\epsilon$ & $\mathbb{N}$ & superpixel scale \\
% $w^{\epsilon}$ & $\mathbb{Z}$ & Weight of scale $\epsilon$ \\
 \bottomrule
\end{tabular}
\end{center}
\end{table}

\subsection{Superpixel Graph Construction}
\label{sec:superpixelgrouping}

Graph-based segmentation methods \cite{shi2000normalized,felzenszwalb2004efficient} integrate the information of both color features and spatial position of pixels (or superpixels) in the procedure of graph construction.
This gives graph-based methods the ability to segment images accurately and keep the spatial continuity of segments in the meantime.
Among the graph-based methods, SE-based methods give a high-quality graph partition by SE minimization.
Furthermore, SE-based methods adaptively determine the number of modules in the partition, which are segments corresponding to sub-regions in images in the case of skin lesion segmentation.
This gives SE-based methods the advantage of accurate segmentation of pixels at lesion borders.

We construct a graph $G=(V, E,\mathbf{W})$ based on the image, where the nodes in $V=\{ v_1,...,v_N \}$ correspond to superpixels of image segmented by SLIC algorithm \cite{achanta2012slic}, the edges in $E$ connect spatially close superpixels, and the weight matrix $\mathbf{W}$ measures the similarities among superpixels.
To utilize the spatial information of superpixels, we only connect the nodes whose spatial distance is smaller than a threshold $r$.
For nodes $v_i$ and $v_j$ with edge connected, the similarity weight $\mathbf{W}_{i,j}$ is computed as follows:
\begin{equation}
    \mathbf{W}_{i,j} = {\rm exp}\left( \frac{-d( v_i, v_j )^2}{\sigma_i \sigma_j} \right),
\end{equation}
where $d(v_i,v_j)$ is the mean color distance between $v_i$ and $v_j$, $\sigma_i$ and $\sigma_j$ are the local scaling parameters for nodes \cite{zelnik2004self}.
We then sparsify this graph into a $K$-nearest neighbor graph by retaining $K$ most significant edges for each node, which avoids the complex dense graphs for subsequent SE minimization.

\begin{figure}[htbp]
\centering
\includegraphics[width=1.0\linewidth]{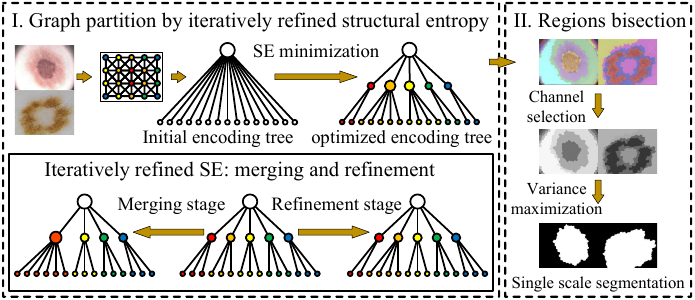}
\caption{Illustration of SE guided segmentation.}
\label{fig:graph_partition}
\end{figure}

\subsection{Structural Entropy Guided Segmentation}
\label{sec:sedguided}
After superpixel graphs construction, images are segmented into regions through SE minimization on these graphs.
The procedure of SE guided segmentation is depicted in Figure \ref{fig:graph_partition}.
We aim to partition the graph into several modules by SE minimization. 
Since nodes in modules represent the superpixels of the image, those modules thus represent homogeneous regions of the image. 
As such, we can simply classify those regions to be either lesion regions or healthy skin regions to predict accurate lesion boundaries. 

We minimize the SE of $G$ to obtain a high-quality graph partition $\mathcal{P}=\{X_1,X_2,...,X_L\}$.
Greedily applying the merging operator \cite{li2016structural} on the encoding tree gives the SE minimized encoding tree and corresponding graph partition with adaptively determined $L$ modules.
However, the partition result by greedy merging can be sub-optimum.
We propose the iteratively refined structural entropy for better results.
It composes a merging stage and a refinement stage, as illustrated in Algorithm \ref{algo:iter}.
In the merging stage, we merge nodes to maximize the decreasing amount of SE to obtain the encoding tree with multiple modules.
And then in the refinement stage, we greedily adjust nodes into different modules to achieve better SE scores.

\renewcommand{\algorithmicrequire}{\textbf{Input:}}
\renewcommand{\algorithmicensure}{\textbf{Output:}}
\begin{algorithm}[htbp]
\caption{Iteratively refined structural entropy}
\label{algo:iter}
\begin{algorithmic}[1]
    \REQUIRE A graph $G=(V,E,\mathbf{W})$
    \ENSURE Encoding tree $\mathcal{T}$ and corresponding partition $\mathcal{P}$
    \STATE Initialize $\mathcal{T}$ containing all graph nodes as tree leaves
    % \STATE // Encoding tree by merging
    \STATE // Merging stage
    \REPEAT
        \STATE Merge a chosen module pair $(X,Y)$ into $X \cup Y$ condition on $argmax_{X,Y} \{ \Delta \mathcal{H}^\mathcal{M}_{X,Y} \}$ via Eq. (\ref{eq:merging})
        \STATE Update $\Delta \mathcal{H}^\mathcal{M}$ for node pairs connected to $X$ or $Y$
    \UNTIL $\Delta \mathcal{H}^\mathcal{M} < 0$ for all module pairs
    % \STATE // Encoding tree by refinement
    \STATE // Refinement stage
    \REPEAT
        \FOR{each node $v_i \in V$}
        \STATE Remove node $v_i$ from the original module $X$
        \STATE Insert node $v_i$ into a chosen module $Y$ condition on $argmax_{Y}\{ \Delta \mathcal{H}^\mathcal{R}_{X,v_i} - \Delta \mathcal{H}^\mathcal{I}_{Y,v_i} \}$ via Eqs. (\ref{eq:remove}) and (\ref{eq:insert})
        \STATE Update the cut and the volume of each module
        \ENDFOR
    \UNTIL $\mathcal{H}^\mathcal{P}(G)$ converges
\end{algorithmic}
\end{algorithm}

In the merging stage, we apply the merging operator on the initial encoding tree which contains all graph nodes as tree leaves to obtain the encoding tree by merging.
For an encoding tree $\mathcal{T}$ with modules $X$ and $Y$ contained, the decrease amount of SE after merging $X$ and $Y$ is given by:
\begin{equation}
\label{eq:merging}
\begin{split}
    \Delta\mathcal{H}_{X,Y}^{\mathcal{M}} = 
    \frac{1}{vol(G)} [ \left( vol(X)-cut(X) \right) {\rm log_2} \, vol(X) \\
    + \left( vol(Y)-cut(Y) \right) {\rm log_2} \, vol(Y) \\
    - \left( vol(X \cup Y)-cut(X \cup Y) \right) {\rm log_2} \, vol(X \cup Y) \\
    + \left( cut(X)+cut(Y)-cut(X \cup Y) \right) {\rm log_2} \, vol(G) ],
\end{split}
\end{equation}
where $\mathcal{M}$ denotes modules $\mathcal{M}$erging, $X \cup Y$ is a module merged from $X$ and $Y$, containing nodes from either $X$ or $Y$, $vol(X)$ and $vol(G)$ are the volume of $X$ and $G$, and $cut(X)$ is the cut of $X$, as defined in Section \ref{sec:se}. 
Merging operator merges $X$ and $Y$ into $X \cup Y$ if $\Delta\mathcal{H}_{X,Y}^{\mathcal{M}} > 0$.

In the refinement stage, we iteratively refine the encoding tree from the merging stage to obtain the encoding tree by refinement.
For graph node $v_i$, the decrease amount of SE by removing $v_i$ from the original module $X$ is given by:
\begin{equation}\label{eq:remove}
\begin{split}
    \Delta\mathcal{H}_{X,v_i}^\mathcal{R} = \frac{vol(X)-cut(X)}{vol(G)} {\rm log_2} \frac{vol(X)}{vol(G)} \\
    - \frac{vol(X\backslash\{v_i\}) - cut(X\backslash\{v_i\})}{vol(G)} {\rm log_2} \frac{vol(X\backslash\{v_i\})}{vol(G)},
\end{split}
\end{equation}
% \begin{equation}\label{eq:remove}
% \begin{split}
%     \Delta\mathcal{H}_{X,v_i}^\mathcal{R} = - \frac{vol(X\backslash\{v_i\}) - cut(X\backslash\{v_i\})}{vol(G)} {\rm log_2} \frac{vol(X\backslash\{v_i\})}{vol(G)} \\
%     + \frac{vol(X)-cut(X)}{vol(G)} {\rm log_2} \frac{vol(X)}{vol(G)},
% \end{split}
% \end{equation}
where $\mathcal{R}$ denotes node $\mathcal{R}$emoving, $X\backslash\{v_i\}$ means removing node $v_i$ from module $X$, and the increase amount of SE by inserting node $v_i$ into another module $Y$ is given by:
\begin{equation}\label{eq:insert}
\begin{split}
    \Delta\mathcal{H}_{Y,v_i}^\mathcal{I} = 
    - \frac{vol(Y)-cut(Y)}{vol(G)} {\rm log_2} \frac{vol(Y)}{vol(G)} \\
    + \frac{vol(Y \cup \{v_i\})-cut(Y \cup \{v_i\})}{vol(G)} {\rm log_2} \frac{vol(Y \cup \{v_i\})}{vol(G)},
\end{split}
\end{equation}
where $\mathcal{I}$ denotes node $\mathcal{I}$nserting, $Y \cup \{v_i\}$ denotes inserting node $i$ into module $Y$.
It should be noted that the constructed graph $G$ contains no self-loop, thus $g_i=d_i$ for graph node $v_i$.
In each iteration, we remove nodes from the original modules and find modules that minimize SE the most.

In the merging stage, the merging operator is performed iff $\Delta\mathcal{H}^\mathcal{M}_{X, Y}>0$, so $\mathcal{H}^\mathcal{P}(G)$ decreases after each merging iteration and converges when no such module pair exists.
In the refinement stage, node $v_i$ transfers from module $X$ to a module $Y$ with largest $\Delta \mathcal{H}^\mathcal{R}_{X,v_i} - \Delta \mathcal{H}^\mathcal{I}_{Y,v_i}$ in each iteration, so $\mathcal{H}^\mathcal{P}(G)$ decreases or stays constant after each refinement iteration, and converges when no improvement can be made.

The time complexity of the merging stage is $O(N{\rm log}^2N)$ \cite{li2016structural}.
In the refinement stage, calculating $\Delta\mathcal{H}_{X,v_i}^\mathcal{R}$ and $\Delta\mathcal{H}_{Y,v_i}^\mathcal{I}$ for each node $v_i$ and every possible module $Y$ take time of $O(NL)$ for each iteration.
Also, updating the cut and the volume of each module take time of $O(NL)$ for each iteration.
Thus, the total time complexity of Algorithm \ref{algo:iter} is $O(N{\rm log}^2N+NLT)$, where $N$, $L$, and $T$ denote the number of nodes, modules, and iterations, respectively.

\noindent\textbf{Homogeneous Regions Bisection.}
% \paragraph{Homogeneous Regions Bisection.}
\label{sec:bisection}
Hereafter, we classify those segmented regions into the lesion region and healthy skin.
We adopt the between-class variance maximization method, which is a widely used criterion in skin lesion segmentation \cite{guarracino2018sdi+}.
Specifically, we choose a color channel $c$ from the image, obtain a gray image, and use the intensity of the gray image to classify the regions into classes $C_0$ and $C_1$.
To better separate the regions, the channel $c$ is chosen from the RGB channels with maximum $\sigma^2_B$ after bisection.
For each region $X_i$, the occurrence probability $\omega_{X_i}$ and the mean intensity $\mu_{X_i}$ are represented as:
\begin{equation}
    \omega_{X_i} = \frac{\| X_i \|}{M},
\end{equation}
and
\begin{equation}
    \mu_{X_i} = \frac{\sum_{\text{pixel}_j\in X_i}I_j}{\| X_i \|},
\end{equation}
respectively, where $I_j$ is the intensity of pixel $j$, $\| X_i \|$ is the number of pixels in region $X_i$, and $M$ is the total number of pixels in the image.
The between-class variance of regions is given by:
\begin{equation}
\label{eq:sigma_b}
\begin{split}
    \sigma^2_B &= \omega_0 (\mu_0-\mu_T)^2 + \omega_1 (\mu_1-\mu_T)^2, \\ 
    % &= \omega_0 \omega_1 (\mu_1-\mu_0)^2,
\end{split}
\end{equation}
where $\omega_k$ is the occurrence probability of class $C_k$ defined as:
\begin{equation}
    \omega_k = \sum_{X_i \in C_i} \omega_{X_i},
\end{equation}
$\mu_k$ is the mean intensity of class $C_k$ defined as:
\begin{equation}
    \mu_k = \sum_{X_i \in C_k} \omega_{X_i}\mu_{X_i},
\end{equation}
 and $\mu_T$ is the total mean intensity of $C_0$ and $C_1$, defined as:
\begin{equation}
    \mu_T = \sum_{X_i \in C_{0,1}} \omega_{X_i}\mu_{X_i}.
\end{equation}
In practice, we arrange $\mu_{X_i}$ in ascending order and separate those regions into two classes, \textit{i.e.}, $C_0 = \{X_i \mid \mu_{X_i} \leq \tau\}$ and $C_1 = \{X_i \mid \mu_{X_i} > \tau\}$ using a threshold $\tau$ which gives the greatest $\sigma^2_B$.
The darker class $C_0$ is the lesion region.

\subsection{Multi-Scale Segmentation Mechanism}
\label{sec:multiscale}
Due to the complex condition of different dermoscopic images, mis-segmentation may occur in SE-guided segmentation.
We develop a multi-scale superpixel-based segmentation mechanism to improve the segmentation accuracy.
In fact, we observed features of pixels in healthy skin are consistent. Thus, we view superpixels in healthy skin as normal, while those in the lesion region are outliers. 
Hence, we can distinguish healthy and lesion regions by the outlier detection method.
Intuitively, a higher outlier score of a superpixel reflects a higher possibility to be a lesion region.

\noindent\textbf{Outlier Detection by Isolation Forest.}
% \paragraph{Outlier Detection by Isolation Forest.}
Isolation Forest~(\textit{i}Forest) \cite{liu2012isolation} detects outliers by building an ensemble of binary trees that separate training data points called \textit{i}Trees.
The susceptibility of data points to being outliers is measured by the average traversal path length on \textit{i}Trees.
The average traversal path length-based outlier scores of \textit{i}Forest are well-graded.
These graded scores help mis-segmented superpixels to be reclassified correctly in the following binary segmentation procedure by thresholding.

% Given a dermoscopic image segmentation with the lesion region and healthy skin, we build an ensemble of \textit{i}Trees using superpixels $\{ v_i, i=1...K \}$ in the healthy skin whose attributes are the average RGB colors.
Given a dermoscopic image segmentation with lesion region and healthy skin, we train an \textit{i}Forest model containing a collection of \textit{i}Trees using superpixels $\{ s_i, i=1,...,K \}$ in the healthy skin whose attributes are the average RGB colors.
With this ensemble of \textit{i}Trees, the average path lengths $h(s_i)$ of all superpixels $\{ s_i, i=1,...,N \}$ are counted, and the outlier score $o$ of an superpixel $s_i$ is calculated as:
\begin{equation}
    o(v_i,\psi) = 2^{-\frac{E(h(s_i))}{c(\psi)}},
\end{equation}
% where $E(h(v_i))$ is the average of $h(v_i)$ from a collection of \textit{i}Trees, and $c(\psi)$ is the normalization term.
where $E(h(s_i))$ is the average of $h(s_i)$ from the collection of \textit{i}Trees, $\psi$ is the number of superpixels in healthy skin, and $c(\psi)$ is the normalization term.
The outlier scores of all superpixels are obtained from the \textit{i}Trees, forming a score map $O$ at the size of the image.

\noindent\textbf{Multi-Scale Integration.}
% \paragraph{Multi-Scale Integration.}
We integrate the outlier scores from different superpixel scales and form the multi-scale score map.
The weighted score of pixel $j$ being an outlier is defined as:
\begin{equation}
    O(j) = \frac{\sum_{\epsilon \in \gamma}\left( w^{\epsilon}O^{\epsilon}(j) \right)}{\sum_{\epsilon \in \gamma}(w^{\epsilon})},
\end{equation}
where $O^{\epsilon}(j)$ is the score of pixel $j$ at scale $\epsilon$, $\gamma$ is the number of different scales, and $w^{\epsilon}$ is the weight of scale $\epsilon$.
We use the $\sigma^2_B$ in Eq.~(\ref{eq:sigma_b}) to indicate the confidence of single-scale segmentation where larger $\sigma^2_B$ should be associated with larger weights.
Specifically, the weight of a scale $\epsilon$ is defined as:
\begin{equation}
    w^{\epsilon} = {\rm \exp}\{ \frac{\sigma^2_B(\epsilon) - \min_{\epsilon \in \gamma}\sigma^2_B(\epsilon)}{\min_{\epsilon \in \gamma}\sigma^2_B(\epsilon)} \},
\end{equation}
where $\sigma^2_B(\epsilon)$ is the between-class variance of scale $\epsilon$.

% After the multi-scale score map is obtained, we apply Otsu's method \cite{otsu1979threshold} to threshold this map, the region with the higher score is the lesion region.
% The binary segmentation result is obtained by applying Otsu's method \cite{otsu1979threshold} on the multi-scale score map.
The binary segmentation result is obtained by applying GHT thresholding \cite{barron2020generalization} on the multi-scale score map.
We also perform hole filling to avoid small holes in the lesion part.
Following~\cite{zortea2017simple}, we compute the region score of every connected component which increase with region size and proximity to the image center.
This is achieved by filtering each connected component with a two-dimensional Gaussian filter.
The connected component with the largest region score is selected as the final lesion region.

%% file: 4-Experiments.tex
\section{Experiments and Results}

\begin{table*}[htbp]
\caption{Comparison of unsupervised skin lesion segmentation methods. \textbf{Bold}: the best performance on each metric, \underline{underline}: the second best performance.\label{tab:main}}
\centering
\begin{tabular}{l|ccccc|ccccc|ccccc}
\toprule
\textbf{}   & \multicolumn{5}{c|}{ISIC 2016}  & \multicolumn{5}{c|}{ISIC 2017} & \multicolumn{5}{c}{ISIC 2018} \\ \cline{2-16} 
Method\%    & AC$\uparrow$ & SE$\uparrow$ & SP$\uparrow$ & DI$\uparrow$ & JA$\uparrow$  & AC$\uparrow$ & SE$\uparrow$ & SP$\uparrow$ & DI$\uparrow$ & JA$\uparrow$  & AC$\uparrow$ & SE$\uparrow$ & SP$\uparrow$ & DI$\uparrow$ & JA$\uparrow$ \\ \midrule
UDASOD & 71.15 & 45.44 & 76.57 & 34.50 & 27.96      & 59.07 & 60.07 & 55.99 & 31.19 & 23.70     & 67.87 & 35.92 & 79.60 & 24.51 & 18.86 \\
Sp. Merging & 86.67 & 67.13 & 95.39 & 69.97 & 61.06     & 79.89 & 59.16 & 88.72 & 54.66 & 46.02     & 84.53 & 69.82 & 92.79 & 70.11 & 60.89 \\
DRC         & 83.78 & 68.51 & 97.11 & 66.84 & 53.32     & 83.77 & 70.35 & \textbf{95.58} & 59.11 & 45.40     & 83.85 & 70.00 & \underline{97.00} & 68.71 & 56.05 \\
Saliency-CCE & 85.91 & 74.72 & 95.70 & 72.18 & 60.64    & 83.87 & 74.07 & 92.98 & 61.77 & 49.53     & 85.51 & 77.44 & 94.34 & 72.85 & 61.91 \\
A2S-v2      & 87.51 & \textbf{82.64} & 93.85 & 75.52 & 65.67     & 82.85 & 68.65 & 92.80 & 61.35 & 51.10     & \underline{86.22} & 72.13 & \textbf{97.35} & 75.01 & 65.94 \\ \midrule
SpecWRSC & 84.83 & 65.95 & 92.69 & 68.76 & 58.05    & 83.03 & 68.21 & 89.96 & 61.01 & 50.52     & 81.20 & 69.39 & 86.34 & 68.97 & 58.41 \\
NCut  & 85.24 & 64.86 & 93.54 & 68.78 & 58.14   & 83.79 & 67.12 & 90.90 & 62.13 & 51.84     & 82.48 & 68.12 & 88.31 & 69.11 & 58.78 \\
$K$-means  & 89.24 & 69.88 & 95.76 & 76.13 & 66.37     & 84.90 & 70.77 & 90.64 & 67.78 & 58.11     & 83.70 & 71.73 & 88.09 & 71.46 & 61.57 \\
SGSCN       & 86.61 & 63.36 & 96.59 & 71.08 & 60.24     &  85.13 & 54.98 & \underline{95.56} & 59.95 & 50.04    & 82.30 & 71.17 & 87.36 & 70.97 & 61.81 \\ \midrule
\framework$_{SS}$ & \underline{90.69} & 77.85 & \underline{97.23} & \underline{81.61} & \underline{72.21}    & \underline{86.96} & \underline{75.98} & 92.98 & \underline{70.62} & \underline{60.56} &  \underline{86.22} & \underline{79.77} & 90.52 & \underline{76.81} & \underline{67.98} \\
\framework$_{MS}$ & \textbf{91.82} & \underline{80.83} & \textbf{97.72} & \textbf{84.33} & \textbf{76.02}   & \textbf{88.81} & \textbf{77.96} & 94.61 & \textbf{73.90} & \textbf{64.45}   & \textbf{86.93} & \textbf{80.12} & 91.84 & \textbf{77.68} & \textbf{69.35} \\ \midrule
Improvement & 4.31 & - & 0.61 & 8.81 & 10.35  & 3.68 & 3.89 & - & 12.13 & 13.35   & 0.71 & 2.68 & - & 2.67 & 3.41 \\ \bottomrule
\end{tabular}
\end{table*}

In this section, we first describe our experimental setup, including datasets, evaluation metrics, baselines, and implementation details of \framework. 
Next, we quantitatively evaluate \framework~on four datasets to verify the effectiveness of \framework.
Besides, we perform ablation studies and case studies to analyze \framework~in detail.
At last, we evaluate the efficiency of \framework.
All experiments are conducted to answer the following questions:
\vspace{-0.4em}
\begin{itemize}[leftmargin=*]
    \item \textbf{Q1:} Does \framework~outperform state-of-the-art baselines regarding skin lesion segmentation?
    \item \textbf{Q2:} Dose our outlier detection-based multi-scale segmentation mechanism effectively utilize the consistent healthy skin prior?
    \item \textbf{Q3:} Is \framework~sensitive to major parameters?
    \item \textbf{Q4:} Is \framework~efficient enough for skin lesion segmentation compared to baselines?
\end{itemize}

\subsection{Experimental Setup}
\noindent\textbf{Datasets.}
We evaluate \framework~on four public benchmark datasets: ISIC 2016 \cite{gutman2016skin}, ISIC 2017 \cite{codella2018skin}, ISIC 2018 \cite{codella2019skin}, and PH2 \cite{mendoncca2013ph}.
The first three datasets are provided by the International Skin Imaging Collaboration (ISIC) archive from the ``Skin Lesion Analysis toward Melanoma Detection" challenge hosted by the International Symposium on Biomedical Imaging (ISBI) in 2016, 2017, and 2018 respectively.
We evaluate the performance of different methods on the test sets, which contain 379, 600, and 1000 RGB dermoscopic images respectively at the resolution ranging from 1022 $\times$ 767 to 4288 $\times$ 2848, along with their corresponding ground truth.
The PH2 dataset is another frequently used benchmark, especially for unsupervised skin lesion segmentation.
It contains a total of 200 RGB dermoscopic images at the resolution of 768 $\times$ 560 pixels, along with their pixel-level lesion ground truth annotated by expert dermatologists.

\noindent\textbf{Evaluation Metrics and Baselines.}
We adopt five metrics used in ISIC challenges for evaluating all compared methods, which are pixel-level accuracy (AC), pixel-level sensitivity (SE), pixel-level specificity (SP), Dice Coefficient (DI), and Jaccard Index (JA), as defined in \cite{gutman2016skin}.
We compare \framework~with a variety of baseline methods including unsupervised skin lesion segmentation methods, an unsupervised foreground extraction method, unsupervised salient object detection methods, and several clustering methods.
For unsupervised skin lesion segmentation methods, we consider superpixel merging (Sp. Merging \cite{patino2018automatic}), saliency map thresholding (Saliency-CCE), and deep learning-based clustering (SGSCN \cite{ahn2021spatial}).
For the unsupervised foreground extraction method, we consider Deep Region Competition (DRC \cite{yu2021unsupervised}), where skin lesions are taken as foreground.
For unsupervised salient object detection methods, we consider Unsupervised Domain Adaptive Salient Object Detection (UDASOD \cite{yan2022unsupervised}) and texture-guided saliency distilling (A2S-v2 \cite{zhou2023texture}), where skin lesions are taken as salient objects.
For clustering methods, we consider normalized cut (NCut \cite{shi2000normalized}), $K$-means, and hierarchical clustering (SpecWRSC \cite{laenen2023nearly}).
The number of segments of $K$-means, NCut, and SpecWRSC is set to be three, which has been reported to perform well for skin lesion segmentation \cite{ahn2021spatial}.
To achieve a fair comparison, we perform the same preprocessing, homogeneous region bisection, and postprocessing method for clustering-based methods as \framework.
The details of important methods are as follows:
\vspace{-0.4em}
\begin{itemize}[leftmargin=*]
    \item \textbf{Sp. Merging \cite{patino2018automatic}} segments skin lesions by greedily merging superpixels using the distance of the mean color of each superpixel as the criterion.
    \item \textbf{Saliency-CCE \cite{zhou2023saliency}} utilizes a hand-crafted color feature extractor called color channel volume to obtain the saliency map of skin lesions and binarize this map through adaptive thresholding to obtain lesion masks.
    \item \textbf{SGSCN \cite{ahn2021spatial}} is a spatial guided self-supervised clustering network that groups pixels that are spatially close and have consistent features.
    \item \textbf{DRC \cite{yu2021unsupervised}} is an unsupervised foreground extraction approach by reconciling energy-based prior with generative image modeling in the form of Mixture of Experts.
    % Foregrounds are lesions in the scenario of skin lesion segmentation. 
    \item \textbf{UDASOD \cite{yan2022unsupervised}} is an unsupervised domain adaptive salient object detection method which learns saliency from synthetic but clean labels.
    % Salient objects are lesions in the scenario of skin lesion segmentation.
    \item \textbf{A2S-v2 \cite{zhou2023texture}} is an unsupervised salient object detection method with a confidence-aware saliency distilling strategy and a boundary-aware texture matching strategy.
    % Salient objects are lesions in the scenario of skin lesion segmentation.
    % \item \textbf{NCut \cite{shi2000normalized}} performs spectral clustering using eigenvectors of the graph Laplacian matrix.
    % \item \textbf{$K$-means} divide data points into $K$ clusters by iteratively assigning each data point to the nearest centroid.
    \item \textbf{SpecWRSC \cite{laenen2023nearly}} is an efficient hierarchical clustering algorithm running in nearly-linear time in the input size of the input graph.
    We choose the largest $K$ modules from the hierarchical clustering tree to obtain $K$ clusters.
\end{itemize}

\noindent\textbf{Implementation Details.}
For all datasets, images are resized into $768 \times 560$ pixels, artifacts in dermoscopic images are removed using existing methods, including hairs \cite{lee1997dullrazor}, dark corners \cite{pewton2022dark}, and color charts \cite{jahanifar2018supervised}.
Color constancy \cite{finlayson2004shades} is also performed on images to cope with illumination problems.
We set the parameters of superpixel graph construction empirically, with the spatial distance threshold $r$ of 0.3 times image height and width, the local scaling parameter $\sigma$ of 30, and retained edge number $K$ of 50.
To evaluate the effectiveness of the multi-scale mechanism in \framework~framework, we report both the results of single-scale \framework~(\framework$_{SS}$) and multi-scale \framework~(\framework$_{MS}$), corresponding to \framework~without and with the multi-scale mechanism.
In the reported results, we empirically set the number of superpixels of \framework$_{SS}$ to be 400.
The segmentation results of \framework$_{MS}$ integrate the results of \framework$_{SS}$ with superpixel numbers from 200 to 700 at increments of 50.
All the experiments are conducted on a server with two Intel(R) 2.30 GHz CPUs and 500 GB RAM.

\subsection{Quantitative Results (Q1)}

Table \ref{tab:main} shows the results of unsupervised skin lesion segmentation methods on three ISIC datasets.
Compared to state-of-the-art methods, \framework$_{MS}$~and \framework$_{SS}$~achieve the best and second-best performance on three metrics including AC, DI, and JA on both three datasets.
\framework$_{MS}$ witnesses improvements from 0.61\% to 13.35\% on these three metrics, as shown in the last row of Table \ref{tab:main}.
Since SE and SP are a pair of metrics to measure how well methods can identify true positives and true negatives respectively, \framework$_{MS}$ failed to achieve the best performance on both metrics.
However, \framework$_{MS}$ achieves the best performance on at least one metric on both three datasets and achieves quite well performance on the other metric, demonstrating the superiority of \framework$_{MS}$ on this pair of metrics.
Both \framework$_{SS}$ and \framework$_{MS}$ achieve better performance than all other clustering-based methods, which thanks to their ability to adaptively determine the number of segments of SE.
We can also see that \framework$_{MS}$ achieves better performance than \framework$_{SS}$ on all metrics, which proves the effectiveness of our outlier detection-based multi-scale mechanism.
% The last row of Table \ref{tab:main} shows the improvements of \framework$_{MS}$ to previous methods.

We also evaluate \framework~on the popular PH2 dataset, as shown in Table \ref{tab:ph2}.
\framework$_{MS}$~and \framework$_{SS}$~achieve the best and second-best performance on four metrics and achieve quite high SP scores comparable to baseline methods.
Improvements of \framework$_{MS}$ on this dataset range from 1.20\% to 3.35\% on four metrics, as shown in the last row of Table \ref{tab:ph2}.
Through quantitative comparison with nine baselines on four datasets, \framework~is proven to achieve new state-of-the-art performance among unsupervised skin lesion segmentation methods.

\begin{table}[t]
\caption{Comparison of unsupervised skin lesion segmentation methods on PH2 dataset. \label{tab:ph2}}
\begin{center}
\begin{tabular}{@{}l|ccccc@{}} \toprule
Method\%    & AC$\uparrow$ & SE$\uparrow$ & SP$\uparrow$ & DI$\uparrow$ & JA$\uparrow$  \\ \midrule
UDASOD & 63.18 & 48.22 & 68.03 & 31.56 & 23.68 \\
Sp. Merging & 89.10 & 79.63 & 96.00 & 82.83 & 73.97 \\
DRC         & 82.63 & 69.18 & \textbf{97.65} & 72.41 & 59.76 \\
Saliency-CCE & 84.62 & 78.90 & 93.77 & 76.79 & 65.62 \\
A2S-v2      & 85.66 & 87.56 & 91.37 & 78.21 & 67.15 \\ \midrule
SpecWRSC & 85.96 & 79.08 & 89.74 & 78.12 & 68.54 \\
NCut & 88.15 & 76.16 & 94.94 & 80.17 & 70.51 \\
$K$-means & 91.64 & 83.82 & 95.03 & 86.14 & 77.78 \\
SGSCN       & 91.80 & 84.26 & 93.82 & 87.89 & \underline{80.15} \\ \midrule
\framework$_{SS}$ & \underline{92.00} & \underline{89.35} & 94.76 & \underline{88.02} & 80.11 \\
\framework$_{MS}$ & \textbf{93.00} & \textbf{89.97} & \underline{96.37} & \textbf{90.34} & \textbf{83.50} \\    \midrule
Improvement   & 1.20 & 2.41 & - & 2.45 & 3.35 \\ \bottomrule
\end{tabular}
\end{center}
\end{table}

\begin{figure*}[ht]
\centering
\includegraphics[width=0.9\linewidth, trim={0 6 0 3},clip]{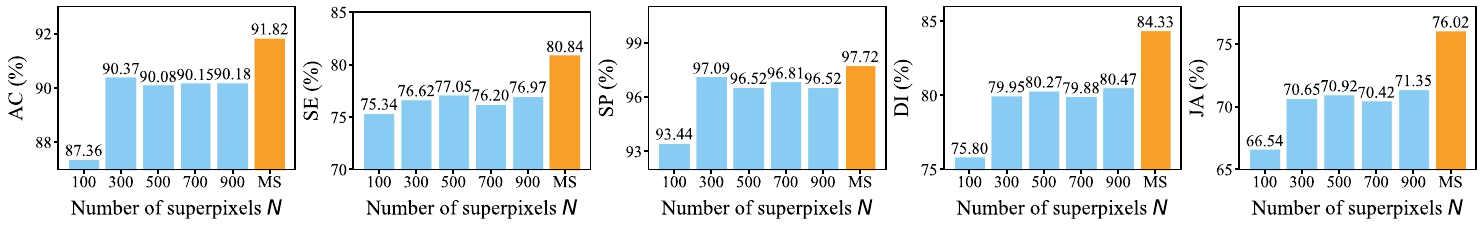}
\caption{Performance of \framework$_{SS}$ with different number of superpixels (blue) and \framework$_{MS}$ (orange) on the ISIC 2016 dataset.}
\label{fig:spnumber}
\end{figure*}

\begin{figure*}[ht]
\centering
\includegraphics[width=0.9\linewidth, trim={0 6 0 3},clip]{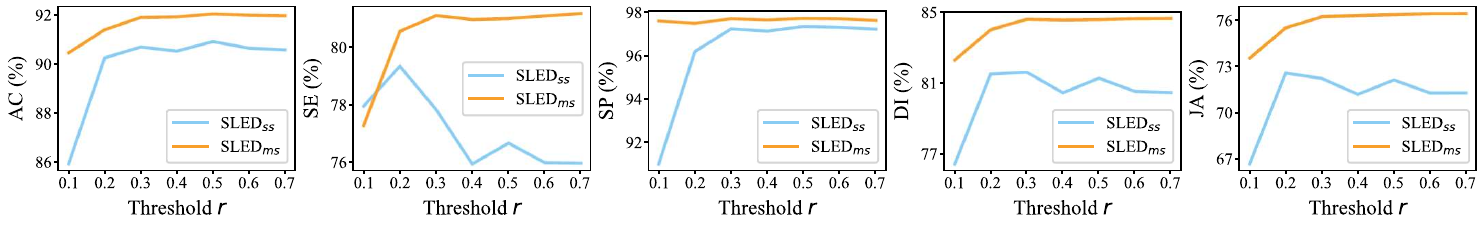}
\caption{Parameter sensitivity experiments for threshold $r$ on the ISIC 2016 dataset.}
\label{fig:thresholdr}
\end{figure*}

\begin{table}[t]
\caption{Performance of \framework~on the ISIC 2016 dataset using different outlier detection methods.\label{tab:outlier}}
\begin{center}
\begin{tabular}{@{}l|ccccc@{}} \toprule
Method\%    & AC$\uparrow$ & SE$\uparrow$ & SP$\uparrow$ & DI$\uparrow$ & JA$\uparrow$  \\ \midrule
\framework$_{Ecod}$   & 75.58 & 24.51 & \underline{99.51} & 33.36 & 23.77 \\
\framework$_{K-NN}$    & 84.80 & 47.94 & \textbf{99.78} & 60.07 & 47.72 \\
\framework$_{OCSVM}$ & \underline{91.34} & \underline{79.52} & 97.48 & \underline{82.59} & \underline{74.13} \\ 
\framework$_{\textit{i}Forest}$ & \textbf{91.82} & \textbf{80.83} & 97.72 & \textbf{84.33} & \textbf{76.02} \\ \bottomrule
\end{tabular}
\end{center}
\end{table}

\begin{figure}[t]
\centering
\includegraphics[width=0.85\linewidth]{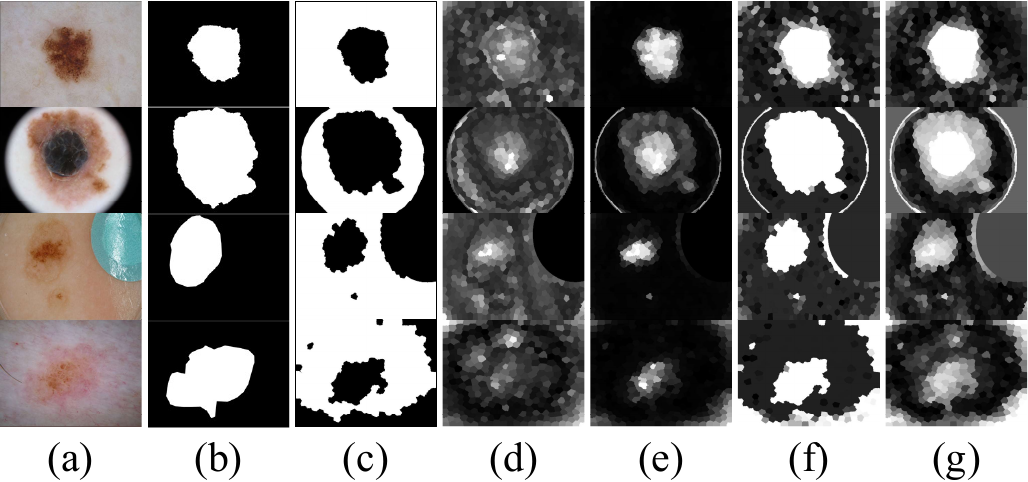}
\caption{Visual comparison of outlier score maps from different outlier detection methods. (a) Dermoscopic images; (b) Ground truth; (c) Detected healthy skin by \framework$_{SS}$; (d)-(g) Outlier score maps from healthy skin by (d) Ecod \cite{li2022ecod}, (e) $K$-NN \cite{angiulli2002fast}, (f) OCSVM \cite{scholkopf2001estimating}, and (g) \textit{i}Forest \cite{liu2012isolation} respectively.}
\label{fig:odss}
\end{figure}

\subsection{Ablation Study (Q2, Q3)}

\noindent\textbf{Sensitivity Analysis.}
We analyze the sensitivity of parameters involved in \framework.
% \framework~involves several parameters which influence the performance of \framework.
One important parameter is the superpixel number $N$.
A larger value of $N$ reduces the risk of over-merge but impeded the efficiency of \framework.
We explore the performance of \framework$_{SS}$ with different values of $N$ as shown in Figure \ref{fig:spnumber}.
The performance of \framework$_{SS}$ is slightly fluctuating but generally stable when $N>100$, indicating that \framework$_{SS}$ adapts well among a wide range value of $N$.
\framework$_{MS}$ outperforms any \framework$_{SS}$ results in four out of five metrics by integrating segmentation of ~\framework$_{SS}$ at different superpixel scales, demonstrating the effectiveness of the multi-scale segmentation mechanism.

Another important parameter is the spatial distance threshold $r$, which is used to balance the importance of color features and the spatial position of superpixels in the graph construction.
A larger value of $r$ means greater importance of the color feature.
Figure \ref{fig:thresholdr} shows the sensitivity of $r$ to \framework.
It can be observed that the performance of \framework~is poor when $r=0.1$ and gets stable when $r$ is larger.
\framework$_{SS}$ witnesses a small drop in performance when $r$ is larger than 0.3, but this drop can be eliminated by the multi-scale segmentation mechanism of \framework$_{MS}$.

\noindent\textbf{Outlier Detection Methods.}
The proposed multi-scale segmentation mechanism is a general approach, which can be implemented by many unsupervised outlier detection methods.
To find the method that best models healthy skin, we utilize different outlier detection methods to implement several variants of~\framework~, including~\framework$_{Ecod}$ (Ecod \cite{li2022ecod}),~\framework$_{K-NN}$ ($K$-Nearest Neighbors \cite{angiulli2002fast}),~\framework~$_{OCSVM}$ (OCSVM \cite{scholkopf2001estimating}), and~\framework~$_{\textit{i}Forest}$ (Isolation Forest \cite{liu2012isolation}).
Table \ref{tab:outlier} presents the performance of these variants.
~\framework~$_{\textit{i}Forest}$ achieves best performance in all metrics, which indicates that \textit{i}Forest models healthy skin better.
~\framework~$_{OCSVM}$ ranks second with comparable performance.

Figure~\ref{fig:odss} illustrates examples of outlier score maps using different outlier detection methods.
Both Ecod and $K$-NN fail to distinguish the lesion region and healthy skin.
OCSVM distinguishes the lesion region but fails to reflect the diverse pixel feature in the lesion region, as shown in the second and third rows.
Visually the outlier score maps of \textit{i}Forest better reflect the features of the lesion region by giving different superpixels graded scores.
% prominent lesion region has higher scores.
When the detected healthy skin by \framework$_{SS}$ is not accurate, as shown by the last row, all methods fail to give a good score map.
However, the score map of \textit{i}Forest still reflects the lesion region.
In all, \textit{i}Forest best models healthy skin, and best fits our framework.

\subsection{Case Study}

\begin{figure*}[t]
\centering
\includegraphics[width=0.9\linewidth]{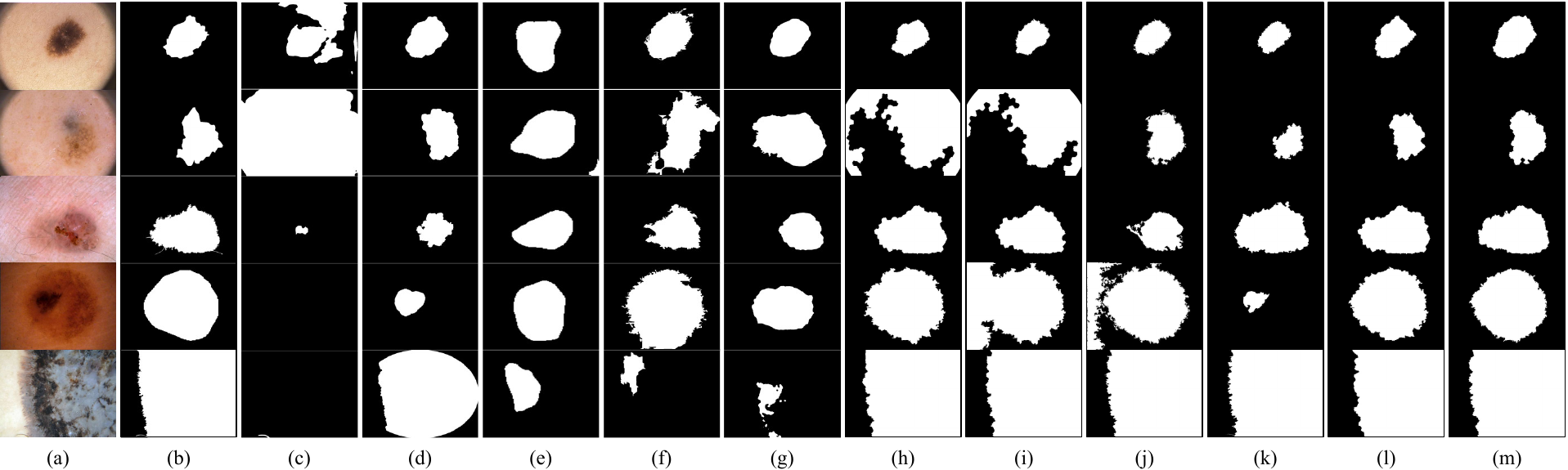}
\caption{Comparison of segmentation results. (a) Dermoscopic images; (b) Ground truth; (c) UDASOD (d) Sp. Merging; (e) DRC; (f) Saliency-CCE; (g) A2S-v2; (h) SpecWRSC; (i) NCut; (j) $K$-means; (k) SGSCN; (l) \framework$_{SS}$; (m) \framework$_{MS}$.}
\label{fig:casestudy}
\end{figure*}

\noindent\textbf{Qualitative Comparison.}
% \paragraph{Qualitative Comparison}
Figure \ref{fig:casestudy} gives a qualitative comparison of different segmentation methods.
Variations of lesions in color, texture, shape, size, and location impede lesion segmentation accuracy.
Low contrast between lesions and healthy skin (shown by 2nd and 3rd rows) and changing illumination (shown by 4th row) are the other two challenges.
Furthermore, many lesions show diverse features and can be divided into several sub-regions, misleading some methods to take less prominent parts of lesions as healthy skin.
Both K-means, NCut, and SpecWRSC suffer from the problem of fixed segments number, leading to poor segmentation results in some cases.
Compared to other methods including \framework$_{SS}$, \framework$_{MS}$ gives more accurate lesion boundaries.

\begin{figure}[t]
\centering
\includegraphics[width=0.85\linewidth]{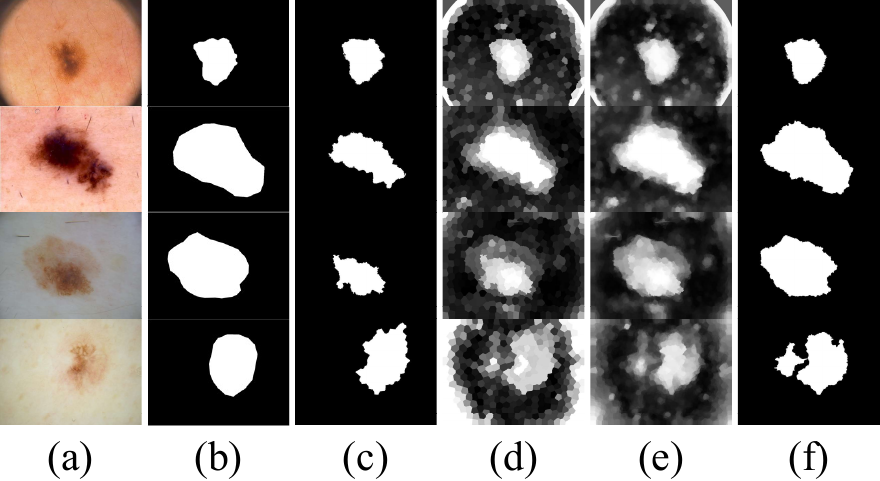}
\caption{Visualization of outlier score maps. (a) Dermoscopic images; (b) Ground truth; (c) Segmentation results of \framework$_{SS}$; (d) Single-scale outlier score maps from (c); (e) Multi-scale outlier score maps; (f) Segmentation results of \framework$_{MS}$ from (e).}
\label{fig:scoremaps}
\end{figure}

\noindent\textbf{Examples of Outlier Score Maps.}
Figure \ref{fig:scoremaps} shows several examples of outlier score maps.
The multi-scale mechanism helps reduce the mis-segmentation of \framework$_{SS}$, as shown in the second and third rows.
 In these two rows, the segmented lesion regions by \framework$_{SS}$ are smaller than the ground truths, leading to inaccurate single-scale score maps.
Multi-scale score maps however better distinguish lesion regions by integrating different single-scale score maps, leading to more accurate segmentation results of \framework$_{MS}$.
It should also be noted that when the features of pixels in healthy skin are not consistent, the proposed multi-scale mechanism induces noises.
For example, the image in the last row has inconsistent backgrounds due to the influence of light illumination, and the segmentation result of \framework$_{MS}$ is worse than \framework$_{SS}$.

\begin{table}[t]
\caption{Time usage of methods on PH2 dataset.\label{tab:time_usage}}
\centering
\resizebox{0.48\textwidth}{!}{
\begin{tabular}{@{}c|ccccccccccc@{}} \toprule
\rotatebox{90}{Method} & \rotatebox{90}{\makecell[c]{UDASOD}} & \rotatebox{90}{\makecell[c]{Sp. Merging}} & \rotatebox{90}{DRC} & \rotatebox{90}{Saliency-CCE} & \rotatebox{90}{A2S-v2} & \rotatebox{90}{SpecWRSC} & \rotatebox{90}{NCut} & \rotatebox{90}{$K$-means} & \rotatebox{90}{SGSCN} & \rotatebox{90}{\framework$_{SS}$} & \rotatebox{90}{\framework$_{MS}$} \\ \midrule
\makecell[c]{Time\\(min)} & 1 & 60 & 292 & 18 & 1 & 25 & 24 & 22 & 396 & 25 & 51 \\ \bottomrule
% \makecell[c]{Time \\(min)} & 59.8 & - & 17.6 & - & 16.7 & 14.8 & 395.7 & 17.9 & 51.1 \\ \bottomrule
\end{tabular}
}
\end{table}

\subsection{Efficiency (Q4)}
We test the efficiency of different methods on the PH2 dataset as shown in Table \ref{tab:time_usage}.
The running time of \framework$_{SS}$ is comparable to Saliency-CCE, NCut, and $K$-means.
By calculating the segmentation of \framework$_{SS}$ with different scales in parallel, the running time of \framework$_{MS}$ is comparable to Sp. Merging, and less than DRC and SGSCN.
Although UDASOD and A2S-v2 consume the least running time, they both require tremendous model training time.
In all, \framework~is quite efficient.

%% file: 5-Conclusion.tex
\section{Conclusion}
In this paper, we have presented \framework, a novel unsupervised skin lesion segmentation framework.
To address the challenges of diverse pixel features within the lesion region and the consistency of pixel features in healthy skin, we introduce two key components in our framework: SE guided segmentation and a multi-scale mechanism utilizing isolation forest outlier detection.
Our extensive experimental evaluations on four widely used skin lesion segmentation benchmarks confirm the superiority of \framework~over existing methods in terms of effectiveness and efficiency.
We believe that \framework~holds great potential for improving dermatological diagnosis and aiding in early disease detection.